\documentclass[letterpaper, 10 pt, conference]{ieeeconf}  

\IEEEoverridecommandlockouts                              

\usepackage{hyperref} 
\usepackage{romannum} 
\usepackage{graphicx} 
\usepackage{subfigure} 
\usepackage{amsmath,mathtools} 
\usepackage{xfrac}
\usepackage{multirow}
\usepackage{booktabs}
\usepackage{fancyhdr} 

\title{\LARGE \bf
Gaze-based Human-Robot Interaction System for Infrastructure Inspections
}

\author{
Sunwoong Choi$^{1}$, Zaid Abbas Al-Sabbag$^{2}$, Sriram Narasimhan$^{1}$, and Chul Min Yeum$^{2}$ 
\thanks{*Partial support for this work was provided through UCLA through their start-up grant to Narasimhan.}
\thanks{$^{1}$Sunwoong Choi is with the Department of Mechanical and Aerospace Engineering and Sriram Narasimhan is with the Department of Civil and Environmental Engineering, University of California, Los Angeles, CA, USA.
        {\tt\small swchoi0501@ucla.edu, snarasim@ucla.edu}}%
\thanks{$^{2}$Zaid Abbas Al-Sabbag and Chul Min Yeum are with the Department of Civil and Environmental Engineering, University of Waterloo, ON, Canada.
        {\tt\small zaalsabbag@uwaterloo.ca, cmyeum@uwaterloo.ca}}%
}

\begin{document}

\maketitle
\thispagestyle{fancy} 
\pagestyle{empty}

\begin{abstract}

Routine inspections for critical infrastructures such as bridges are required in most jurisdictions worldwide. Such routine inspections are largely visual in nature, which are qualitative, subjective, and not repeatable. Although robotic infrastructure inspections address such limitations, they cannot replace the superior ability of experts to make decisions in complex situations, thus making human-robot interaction systems a promising technology. This study presents a novel gaze-based human-robot interaction system, designed to augment the visual inspection performance through mixed reality. Through holograms from a mixed reality device, gaze can be utilized effectively to estimate the properties of the defect in real-time. Additionally, inspectors can monitor the inspection progress online, which enhances the speed of the entire inspection process. Limited controlled experiments demonstrate its effectiveness across various users and defect types. To our knowledge, this is the first demonstration of the real-time application of eye gaze in civil infrastructure inspections.

\end{abstract}

\section{INTRODUCTION}

Critical infrastructures such as bridges worldwide are deteriorating at an alarming rate, with the American Society of Civil Engineers estimating that 7.5\% of the bridges in the United States are in poor condition. Routine inspections are crucial (and required) to identify potential structural weaknesses and prioritize repair decisions~\cite{ASCE_2021}. Visual inspections are by far the most common method employed to identify potential structural defects such as cracks, spalling, or corrosion, which are visible on its surface~\cite{mto_2008,AASHTO_2019}. However, visual inspections tend to be qualitative, subjective, and sometimes unreliable \cite{c_tmp_intro4}.

Robotic infrastructure inspections have seen increased adoption as they can make inspections more quantitative, reliable, and repeatable \cite{chen2019uav,feroz2021uav,c_tmp_intro1}. However, such systems, especially when implemented autonomously, are unable to prioritize defects, face quality issues in poor lighting conditions, and are unable to generalize to previously untrained defects. For these reasons, human-robot interaction (HRI) systems that can harness domain experts' experience and knowledge to handle unforeseen situations are seen as promising technologies for infrastructure inspections \cite{liang2021human}. 

\begin{figure}[t]
\centerline{\includegraphics[width=0.95\linewidth]{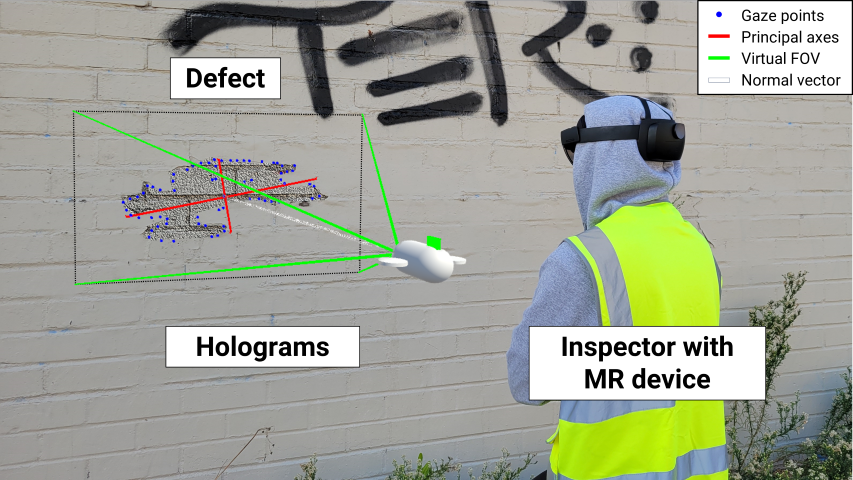}}
\caption{An example of gaze-based human-robot collaborative inspection through mixed reality.}
\label{fig_example}
\end{figure}

Mixed Reality (MR) is an emerging technology merging the physical and virtual worlds by displaying interactive computer graphics \cite{c_tmp_intro2}. Unlike remote control operations for the HRI systems, which require skilled inspectors who can control robotic systems \cite{c_tmp_intro4}, head-mounted MR devices are easier tools for infrastructure inspection by allowing users to interact with the system in an intuitive manner, such as hands, voices, and eye-gaze \cite{c_tmp_intro8}. Additionally, they can visualize data in real time by augmenting holograms to the real objects without constraints of movement \cite{c_tmp_intro2}. Therefore, many MR technology inspection applications have been developed to assist inspectors in decision-making \cite{c_tmp_intro2}.

Mascarenas et al. \cite{c_tmp_intro10} employed an MR device for area measurements of irregular geometric structures. The interesting region was selected by manually using their hands, and then the area of the site was calculated from polygons provided by the MR device. Al-Sabbag et al. and Karaaslan et al. \cite{c_tmp_intro9,c_tmp_intro12} combined the vision-based deep neural network and the inspector's judgment to segment defects. The damage segmentation process was performed interactively with the user so that it could make a correction when the algorithm showed poor quality. While the use of MR technology has generated considerable interest for inspection applications, they require inspectors' intervention during the inspection process for defect quantification and inference. 

Eye gaze inputs \cite{c_tmp_intro11} have tremendous potential in inspection applications and can complement infrastructure inspections by providing vital context and disambiguating visual data. Moreover, the gaze is a fast, hands-free method that requires relatively little training \cite{c_tmp_intro3} and can be gathered while the activity is performed non-intrusively. Nevertheless, to our knowledge, no published study harnessing gaze data in real time for infrastructure inspections exists. Gaze information can also guide control strategies for fully autonomous HRI inspection systems, such as those that can mimic expert inspectors. However, using gaze-based inputs is fraught with challenges, for example, some eye movements are non-intentional \cite{c_tmp_intro3}, or can be difficult to analyze in uncontrolled and dynamic environments, outside controlled laboratory settings.

{\em Contributions:} We propose a novel gaze-based HRI system for infrastructure inspections using an MR device (Fig. \ref{fig_example}). As the first task, we develop a method to categorize human attention into multiple levels based on gaze trajectory and to dismiss meaningless eye movements. Next, a holographic drone (MR drone) is displayed via a head-mounted MR device to interact with an inspector during the inspection non-intrusively. Finally, the gaze trajectory collected while inspecting a defect is evaluated for visual data collection, specifically defect quantification. We evaluated our system using limited controlled experiments in the laboratory.

\section{METHODOLOGY}

\subsection{Overview and architecture}
\begin{figure}[t]
\centerline{\includegraphics[width=\linewidth]{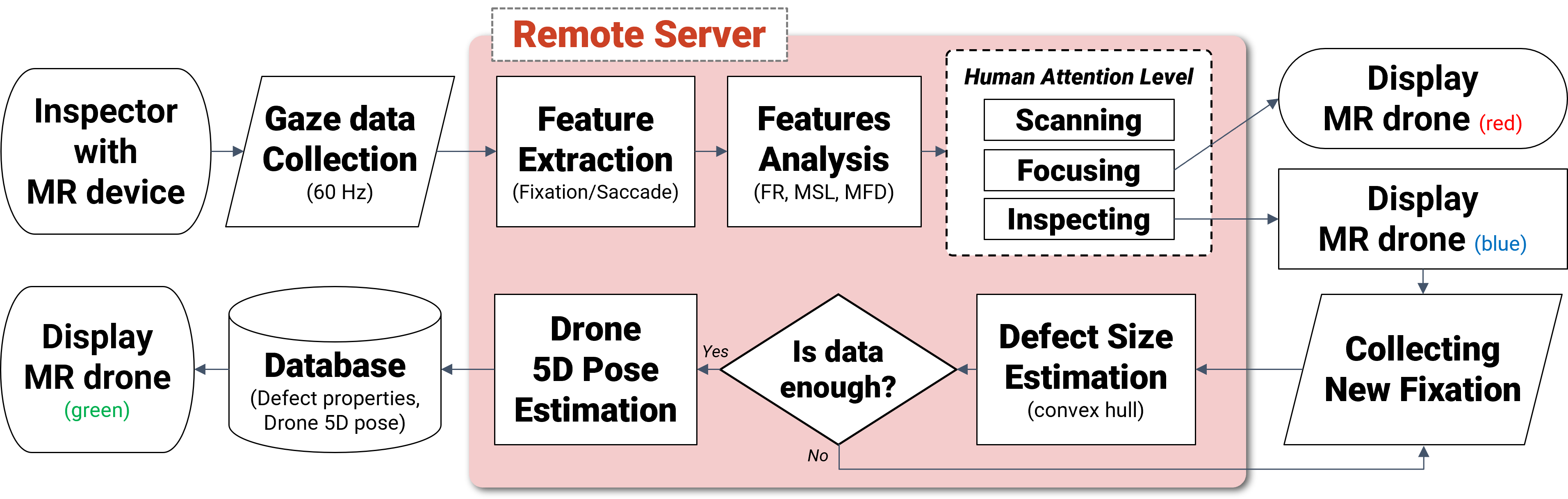}}
\caption{Architecture of the proposed gaze-based inspection system.}
\label{fig_overview}
\end{figure}

Our main contributions in this paper are best described within the context of an inspection system (Fig. \ref{fig_overview}). This system uses gaze data from an MR device as an input to extract gaze features. These features are then analyzed to determine the human attention level: scanning, focusing, or inspecting. When the human attention level is categorized as inspecting, the gaze trajectory is saved to estimate the defect's size and normal vector. Finally, based on this information, the position and orientation of the robot (drone) for visual data collection are determined. HoloLens 2 (HL2), the MR device from Microsoft, is used to collect the raw eye gaze data from the user. The collected gaze data is transferred to the remote server developed by ROS 1. After the process is complete, the results are transferred to the HL2 to display the results, particularly by the MR drone, allowing the user to track the inspection progress.

\subsection{Gaze feature extraction}
Eye movements for visual processing of the environment consist of two main events: fixations and saccades \cite{c_tmp_method1}. A fixation is a brief period of time when eyes pause over a specific area to extract visual information and a saccade is a rapid eye movement from one fixation to the next. 
Among the several methods to identify fixations \cite{c_tmp_method1}, we adopt the dispersion-based method, which classifies gaze points as a fixation if they are within the specific size of the window. HL2 can provide the three-dimensional (3D) point of the intersection between the gaze vector and the target object so that the dispersion-based method can easily identify fixations in the 3D space. In the case of the two-dimensional (2D) eye movement trajectory, the diameter of the fixation is typically within 2--3 deg from the eye, and the minimum duration is 100--150 ms \cite{c_tmp_intro3,c_tmp_method2}. This diameter of the circle is used to describe a sphere to identify fixations in the 3D coordinate system. 

In our experiments, raw eye gaze data was collected at 60 Hz using HL2, so 8 or more gaze points (at least 130 ms) had to be inside the sphere of a diameter of 2.86 deg. This fixation identification process was conducted in real time, so every new gaze point was tested in sequence. Once a gaze point was placed outside the sphere, the identification process was stopped and the mean value of the gaze points was categorized as a fixation or a saccade based on the number of gaze points. Then the identification process was started again with the new gaze point. 


\subsection{Human attention level analysis}
Visual search such as an inspection is a perceptual task of finding a target that is a specific feature or object in the environment with distractors \cite{c_tmp_method3}. During an infrastructure inspection, the visual search process can be divided into three steps based on the human attention level: 1) scanning, 2) focusing, and 3) inspecting \cite{c_tmp_method11}. The eye movement indicates the region of interest where the human is focused \cite{c_tmp_method5} and hence the attention level can be quantified by analyzing the resulting eye movements \cite{c_tmp_method4}. 

There are several eye-tracking metrics to analyze eye movements \cite{c_tmp_intro3}. To measure human attention level, fixation rate (FR), mean fixation duration (MFD), and mean saccade length (MSL) are adopted in this study. The reference time of all three metrics can be arbitrarily set (say, 5s for this application).

FR is the proportion of time spent on fixations during a specific period. The number of fixations decreases during scanning, so the total rate of time spent on fixations also decreases. As accurate visual information is only acquired from fixations \cite{c_tmp_method6}, FR indicates the concentration level of the human during the inspection. 

MFD is the average of the time spent on each fixation. The complexity of the environment increases fixation duration \cite{c_tmp_method7}. Furthermore, fixation duration varies depending on the task. For example, previous studies have shown that the fixation duration for visual search is around 180--275 ms while it is extended to 260--330 ms for scene perception \cite{c_tmp_method6}. This shows that there is a distinctive variation in the duration of fixation between when an inspector scans the area and when the inspector carefully observes a defect.

MSL is the average of the length between the two adjacent fixations. Infrastructure inspections involve the evaluation of several individual components (e.g., beams, columns, foundations); hence even small eye movements can significantly change the target and nature of defects. However, defects are typically localized on a plane or a single structure component. Therefore, when inspectors focus on a defect, fixations are assumed to be close in the 3D world. Studies have shown that during the scene perception, the eye moves 4--5 deg for a saccade \cite{c_tmp_method6}, so a target 1 m away translates to MSL of approximately 8.7 cm.

These aforementioned eye-tracking metrics allow human attention levels to be inferred and used as follows. First, scanning an area is assumed to be the baseline activity during an inspection. To find a suspicious area over the wide area of the critical infrastructure, an inspector quickly scans the area, which is considered the default state of human attention level as it constitutes most of the inspection activity. Therefore, the information obtained during the scanning phase is not used beyond situational awareness. 
\begin{table}[t]
\caption{Attention levels used for experiments}
\label{table_criteria}
\begin{center}
\begin{tabular}{c c c c}
\hline
Human Attention Level & FR & MSL & MFD \rule{0pt}{2.4ex} \rule[-1.2ex]{0pt}{0pt} \\
\hline
Scanning & - & - & - \rule{0pt}{2.4ex} \\
Focusing & 50\% & 0.5 m & - \\
Inspecting & 90\% & 0.15 m & 300 ms \rule[-1.2ex]{0pt}{0pt} \\
\hline
\end{tabular}
\end{center}
\end{table}
During the focusing step, an inspector identifies and focuses on an area of the structure likely to contain defects based on either history, experience, or visual clues. When people are assigned a visual search task, their average FR was found to be 48\% \cite{c_tmp_method4}. For example, if we assume that an inspector is located at a distance of 4--5 m from the defect, this translates to MSL of 0.44 m. This value of FR of 50\% 
and MSL of 0.5 m provides reasonable thresholds for triggering the next step.

The final step involves the inspector observing a defect carefully and recording details about it. Generally, FR tends larger during focus on the scene to gather pertinent information \cite{c_tmp_method1,c_tmp_method2}; say, MFD is at least 300 ms as inspecting a defect is generally more involved than the other two steps \cite{c_tmp_method6,c_tmp_method8}. For this study, the criteria for the inspection step are set to FR of 90\%, MFD of 300 ms, and MSL of 0.15 m. The last threshold assumes that an inspector tends to observe identified interest areas and defects at a closer distance than the previous steps. Table \ref{table_criteria} shows the criteria for three human attention levels used in the experiments.


\subsection{Displaying attention levels during inspection}
 Eye gaze inputs are persistent for the entire period the device is on, and a gaze-based system may have to be personalized \cite{c_tmp_intro3}. To accomplish this, an inspector needs situational awareness while using or gathering eye gaze data as input so that the attention level metrics are consistent with their intention. Such situational awareness will allow users to regulate their behavior for infrastructure inspection tasks \cite{c_tmp_method12}. This study uses a holographic display of an MR drone (visual cue) to provide the inspector with this awareness (inspection status). During the scanning step, the visual cue is not displayed as it is the default status of the inspection and can prove distracting to the inspector during the initial scanning phase. An MR drone appears when the user focuses on a specific area. The role of the visual cue in this step is to make the user aware that they are focusing on an area and that they can consciously control eye movements to align with their intention. 

Once a defect is identified and the attention levels for inspection are met, the color of the MR drone is changed to alert the user, and the fixation data are collected. The size and normal vector of the defect are estimated from this data, and the position and orientation (pose) of the MR drone are determined. Finally, the MR drone whose pose is optimized for capturing the defect's visual data (images) is displayed to the inspector. This allows the inspector to confirm the inspection results in real-time and store them for downstream analysis and reporting. 

\begin{figure}[t]
\centerline{
    \subfigure[]{\includegraphics[scale=0.72]{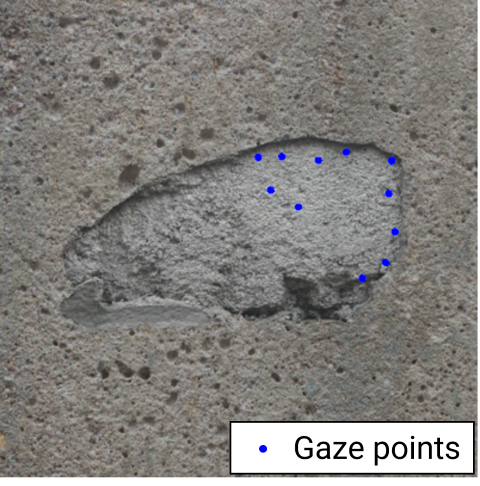}}
    \hfill
    \subfigure[]{\includegraphics[scale=0.72]{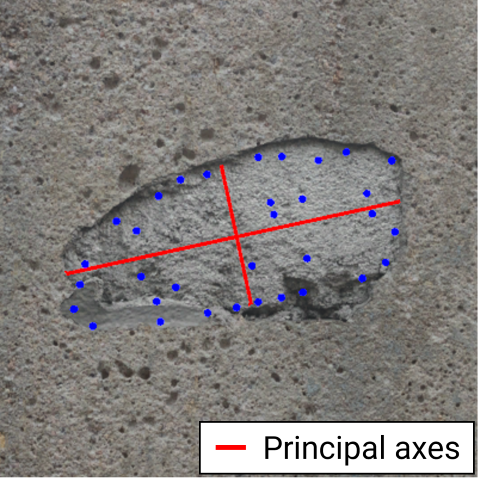}}
    \hfill
    \subfigure[]{\includegraphics[scale=0.72]{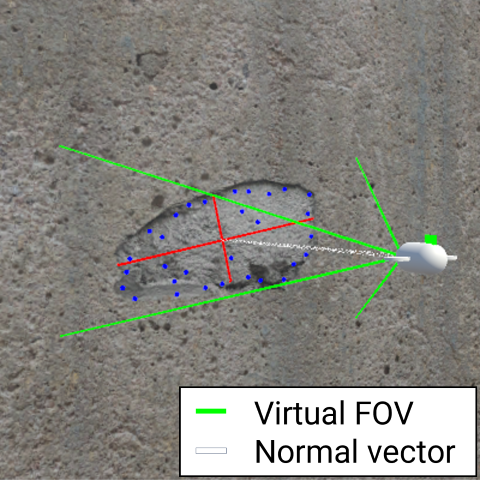}}}
\caption{Analysis of fixations collected during the inspecting step: (a) fixation collection until there is little or no change of the convex hull size of fixations; (b) defect properties evaluation; and (c) 5D pose estimation of the MR drone.}
\label{fig_gazeanalysis}
\end{figure}


\subsection{Pose estimation of the MR drone}
The pose of the MR drone is estimated only when the human attention level reaches the inspecting step, and thus only gaze data collected during this step is analyzed in three stages (Fig. \ref{fig_gazeanalysis}). The MR drone has a total of five degrees of freedom (5D), which can be used to acquire images of the defects. This consists of the 3D position ($x,y,z$) and 2D orientation ($\theta_{pan}$, $\theta_{tilt}$). This study assumes a single-axis tilting camera, allowing the drone's camera to be adjusted to view both upwards and downwards. 

\subsubsection{Defect size estimation}
The first stage for the 5D pose estimation of the drone involves the size estimation of the defect from fixations collected during the inspecting step. As HL2 provides the surface normal vector of the point where the user is looking, the average normal vector for all fixations can be estimated.
Then, the opposite direction of the average normal vector is separated into YX Euler angles on the basis of a right-handed, Y-down coordinate system through inverse kinematics as follows:
\begin{equation}
    \theta_{x} = \arctan2(v_2, \sqrt{1-v_2^2})
\label{eq_IK_x}
\end{equation}
\begin{equation}
    \theta_{y} = \arctan2( \text{-}v_1/\sqrt{1-v_2^2},  \text{-}v_3/\sqrt{1-v_2^2})
\label{eq_IK_y}
\end{equation}
where, $\theta_x$ and $\theta_y$ are the angle about the x-axis and y-axis, respectively, and $[v_1, v_2, v_3]^T$ is the average normal vector of the defect. If $|\theta_x|$ equals to 90 deg, $\theta_y$ is set to be 0 deg.

Fixations in the 3D space can be projected onto the 2D space by rotating $\text{-}\theta_x$ and $\text{-}\theta_y$ and removing the depth information. These transformed fixations represent the fixations from the drone viewpoint. To estimate the defect size, a convex hull of the 2D points is obtained using the Qhull algorithm \cite{c_tmp_method9}. During inspections, inspectors tend to focus on distinctive features such as the boundary of the defect \cite{c_tmp_method11}. Hence, the area of the convex hull of the fixations is assumed to be highly correlated to the size of the defect, which was also verified through our own experiments. Following this, the two principal axes of the convex hull are determined through principal component analysis (PCA) \cite{c_tmp_method10} to calculate the width ($w$) and height ($h$) of the convex hull in the direction of principal axes ($w \geq h$) and the rotation of the longer principal axis based on the z-axis ($\theta_{z}$).

\subsubsection{Data collection time}
The gaze data collection time during the inspecting step correlates with the defect size; the larger a defect is, the greater the data collection time required. For this, the area of the convex hull of the fixations is estimated every time a new fixation is collected. When the inspector observes most of the area of the defect, there is little or no change in the size of the convex hull.
The criterion for data collection time as a function of the convex hull area of the fixations is as follows:
\begin{equation}
    \text{cvx}(p_n) - \frac{\Sigma_{i=n-5}^{n-1}\text{cvx}(p_i)}{5}  \leq \frac{\text{cvx}(p_n)}{100}
\label{eq_time}
\end{equation}
where, $\text{cvx}(p_i)$ is the convex hull area of fixations from the initial fixation $p_0$ to the current fixation $p_i$. Once the Eq. \ref{eq_time} are met, the collection process is stopped, and then all saved fixations are analyzed to determine the 5D pose of the drone.

\subsubsection{5D pose of the MR drone}
Next, following data collection, the drone's 3D position and 2D orientation are determined. There are two types of defects assumed based on where the defect occurs; horizontal (on vertical surfaces such as walls) and vertical (on horizontal surfaces such as ceilings and floors). We use the tilt angle of the average normal vector ($\theta_{x}$) to classify defect type. If the absolute value of $\theta_{x}$ is between 80--90 deg, the defect is considered a vertical case; otherwise horizontal. Based on the defect type, the drone's 2D orientation is as follows:
\begin{equation}
\begin{split}
    &\theta_{tilt} = 
    \begin{dcases}
      \text{sgn}(\theta_{x})\cdot \pi/2 & \text{if}\ |\theta_{x}| \geq 80^{\circ}  \\
      \theta_{x}  & \text{otherwise}
    \end{dcases} \\
    &\theta_{pan} = 
    \begin{dcases}
      \theta_{y} -\text{sgn}(\theta_{x})\cdot\theta_{z} & \text{if}\ |\theta_{x}| \geq 80^{\circ} \\
      \theta_{y}  & \text{otherwise}
    \end{dcases}
\end{split}
\end{equation}

Images used to capture the visual information about the defect are usually in ratios 16:9 or 4:3, which cannot be rotated as the drone pans horizontally (by drone's rotation) and tilts vertically (by the camera's rotation). Therefore, the images should be taken not to exceed the frame with the long axis in the horizontal direction.  The following indicator is used to determine the reference axis between the longer axis and the shorter axis of the photo frame:
\begin{equation}
    \Omega = \frac{W}{H \cdot AR} 
\label{eq_reference}
\end{equation}
\begin{equation}
    W = \max{\{ w\cos(\theta_{z}),h\sin(\theta_{z}) \}},
\end{equation}
\begin{equation}
    H = \max{\{ w\sin(\theta_{z}),h\cos(\theta_{z}) \}},
\end{equation}
where $AR$ is the aspect ratio of a photo, e.g., 1.78 for a 16:9 aspect ratio. In the vertical case, $\theta_{z}$ equals to zero as the drone can rotate horizontally.

When $\Omega$ is equal to or greater than 1, $W$ is the reference length to determine the drone's distance to capture an image, otherwise, $H$ is the reference length. The drone's distance based on the length of the reference axis is as follows:

\begin{equation}
    d = 
    \begin{dcases}
      \frac{W}{2} \tan(\theta_{h}/2) \cdot SF & \text{if}\ \Omega \geq 1 \\
      \frac{H}{2} \tan(\theta_{v}/2) \cdot SF & \text{otherwise}
    \end{dcases}
\end{equation}
where, $\theta_{h}$ and $\theta_{v}$ are the horizontal and vertical fields of view of the camera, respectively, and $SF$ is the safety factor to cover the additional area surrounding the defect, which is set to be 1.5. Finally, the drone's 3D position is given by, 
\begin{equation}
    \boldsymbol{\vec{p}}_{drone} = \boldsymbol{\vec{p}}_{cvx} - \boldsymbol{R_y}(\theta_{pan})\boldsymbol{R_x}(\theta_{tilt})\boldsymbol{\vec{z}}
\end{equation}
where, $\boldsymbol{\vec{p}}_{cvx}$ is the center of the convex hull of the fixations in 3D space, $\boldsymbol{\vec{z}}$ is $[0,0,d]^T$, and $\boldsymbol{R_x}$ and $\boldsymbol{R_y}$ are rotation matrices about x-axis and y-axis, respectively.

\section{EXPERIMENTAL RESULTS}


\subsection{Gaze accuracy evaluation}
\begin{figure}[t]
\centerline{
    \subfigure[]{\includegraphics[scale=0.15]{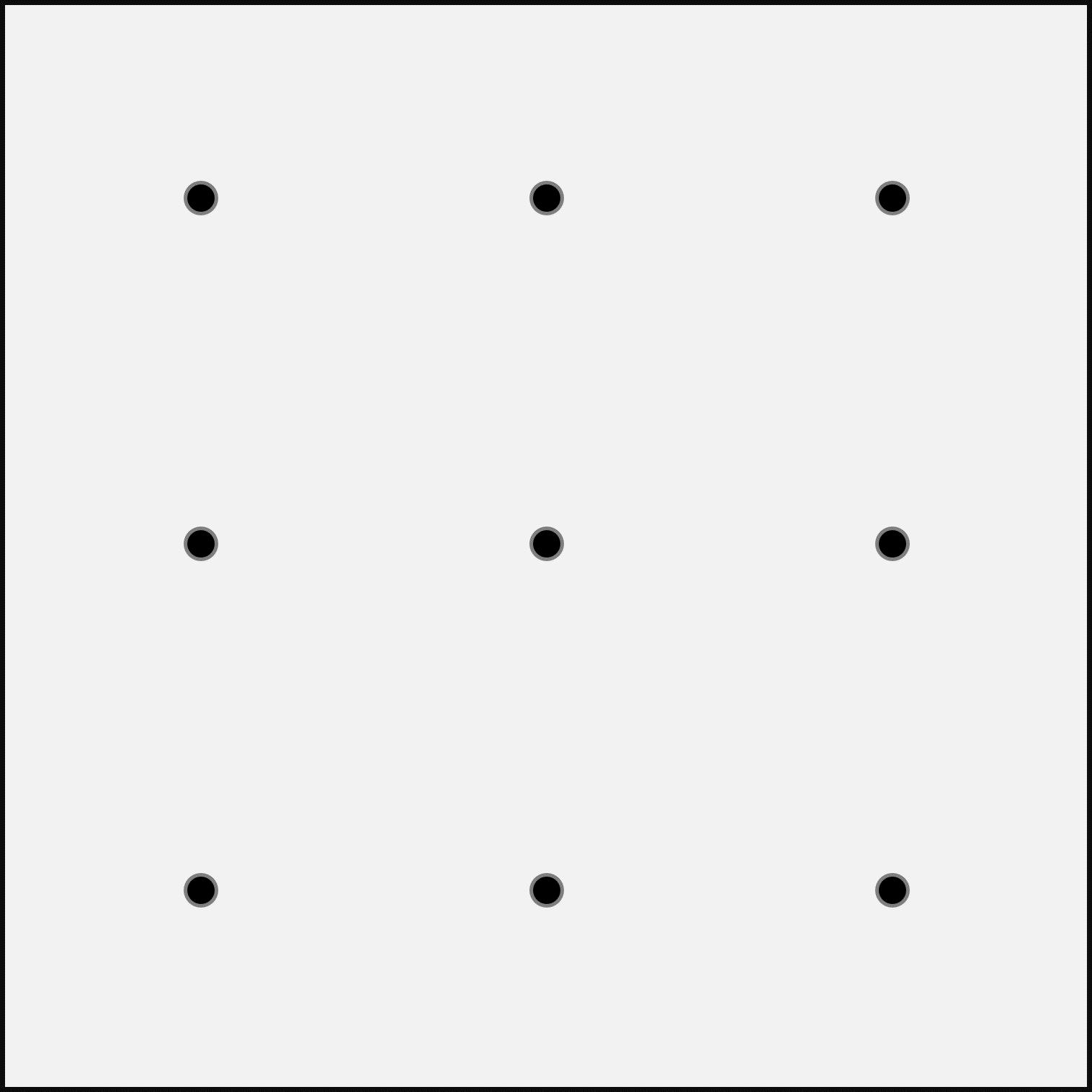}}
    \hspace{8mm}
    \subfigure[]{\includegraphics[scale=0.15]{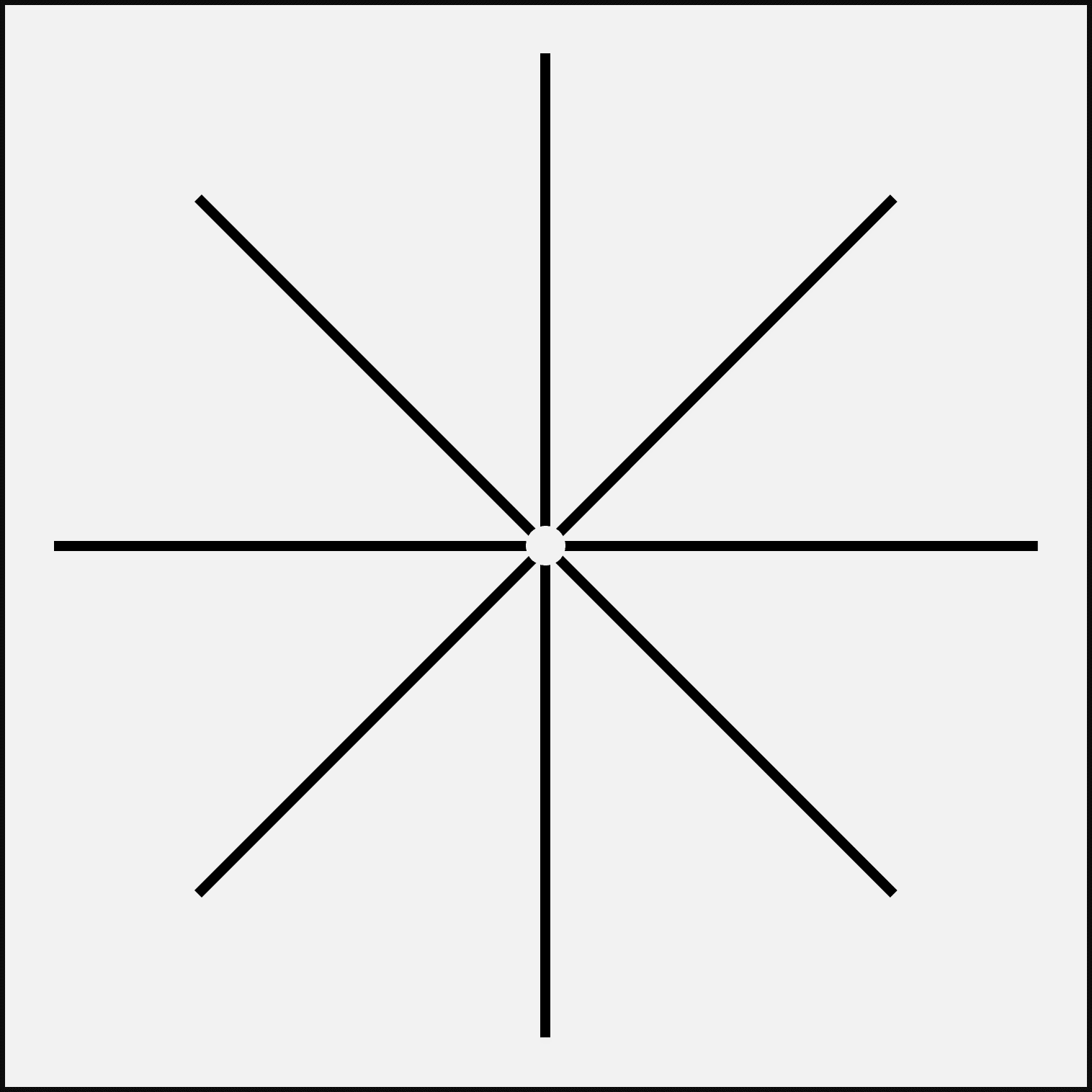}}}
\caption{Eye tracking accuracy test: (a) 9 points for gaze positioning test (b) 8 lines for line tracking test.}
\label{fig_exp1_test}
\end{figure}
Gaze accuracy is closely related to the overall performance of a gaze-based HRI system proposed. In addition to the gaze-pointing ability for the exact location of defects in the infrastructure, the line tracking ability is also crucial as an inspector generally follows distinctive features such as the boundaries of the defect. Gaze positioning accuracy was evaluated using 9 points with an interval distance of 50 cm (Fig. \ref{fig_exp1_test}-a), and line tracking performance was tested using 8 lines of length 70 cm (Fig. \ref{fig_exp1_test}-b). These two targets were placed at different distances, ranging from 1.0 to 6.0 m with an interval of 50 cm considering the limitation of the HL2 sensors. All tests, such as staring at a point or tracking a line, were conducted for 5 seconds.

\begin{figure}[t]
\centerline{\includegraphics[width=1.0\linewidth]{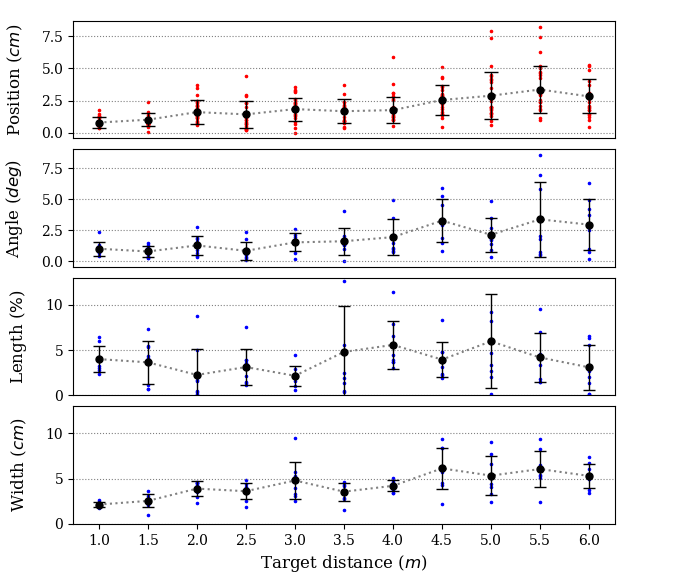}}
\caption{Result of gaze accuracy evaluation; red dots are from the gaze positioning test, and blue dots are from the line tracking test.}
\label{fig_exp1}
\end{figure}

For the gaze positioning accuracy test, the position error of fixations was evaluated (Fig. \ref{fig_exp1}). Generally, 10--11 fixations were generated while looking at a point for 5 seconds, so a total of about 100 fixations were evaluated at each target distance. Position error of fixations had the lowest value of 0.80 cm (SD = 0.41 cm) at the target distance of 1 m, and the highest value of 3.37 cm (SD = 1.85 cm) at the target distance of 5.5 m. The gaze positioning accuracy was decreased when the target was far away. 

The angle, length, and width errors were evaluated to assess the line tracking performance (Fig. \ref{fig_exp1}). The two principal axes of the fixations collected during line tracking were extracted using PCA. The longer axis represents the length of the line, while the shorter one represents the width. The majority of lines had less than an angle error of 2.5 deg and a length error of 5\%, so it can be said that the line tracking accuracy is acceptable for inspection. The width of the lines, which was expected to be close to zero, measured nearly 5 cm. This deviation suggests that defects with a width of less than 5 cm need to be categorized as a crack.

\subsection{Attention level evaluation} 

\begin{figure}[t]
\centerline{\includegraphics[width=0.8\linewidth]{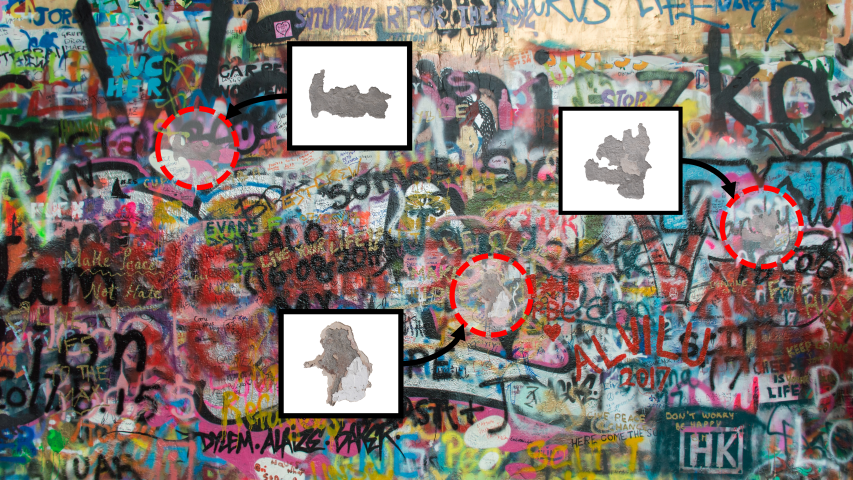}}
\caption{Experimental setup for human attention level evaluation.}
\label{fig_exp3}
\end{figure}

Proper selection of criteria of human attention level is important for the proposed HRI inspection system. A virtual experimental environment was set up as shown in Fig. \ref{fig_exp3}. To mimic the real inspection situation, randomly selected defects (about 10 cm) were placed on the wall (3.5 m $\times$ 2 m) along with many distractors. Participants did not have any prior knowledge about the shape of the defect and their number. Each participant (2 male, 1 female; age 25--35) conducted three sets of visual scanning tasks.

\begin{figure}[t]
\centerline{\includegraphics[width=1.0\linewidth]{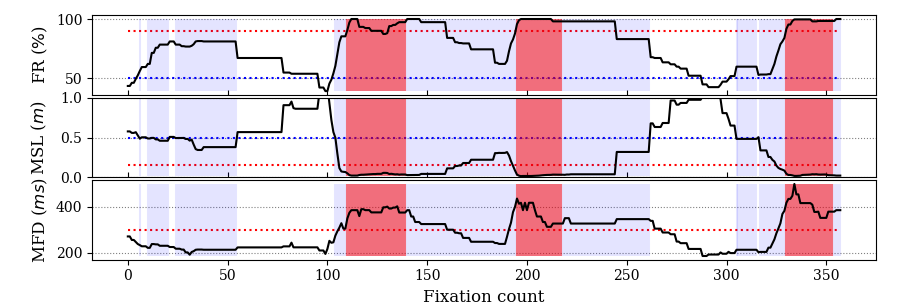}}
\caption{An example of results from a visual searching task; the blue and red dotted lines are the focusing and inspecting step thresholds, respectively. The blue area is the period that the MR drone is displayed and follows the gaze (focusing step). The red area is the data collection period to evaluate the defect (inspecting step).}
\label{fig_exp3_result}
\end{figure}

Fig. \ref{fig_exp3_result} shows an example of the results from the visual scanning task, which included three defects on the wall. Generally, participants entered the focusing step before finding a defect on the wall. Once the defect was detected, FR and MFD were sharply increased and MSL was decreased to enter the inspecting step. Table \ref{table_exp2_result} shows the result of visual searching tasks, where $t_F$ and $t_I$ are the ratios of time duration during the task for the focusing step and the inspecting step, respectively, $t_{avg}$ is the average duration of the three tests, and $n$ is the number of defects on the wall. While all participants spent the majority of their time on the focusing step, they successfully identified the correct number of defects without inadvertently entering the inspecting step. Therefore, this result demonstrates that the human attention level was appropriately selected and the MR drone effectively serves the function of a visual indicator. 

\begin{table}[t]
\caption{The ratio of the focusing and inspecting step during the task}
\label{table_exp2_result}
\begin{center}
\begin{tabular}{c c c c c c c c}
\hline
\multirow{2}{*}[-2pt]{P} & \multicolumn{2}{c}{test1 (n=3)} & \multicolumn{2}{c} {test2 (n=2)} & \multicolumn{2}{c}{test3 (n=1)} & \multirow{2}{*}[-2pt]{$t_{avg}$ (sec)} \rule{0pt}{2.4ex} \\ 
\cmidrule(lr){2-3} \cmidrule(lr){4-5} \cmidrule(lr){6-7}
    & $t_F$ & $t_I$ & $t_F$ & $t_I$ & $t_F$ & $t_I$ \rule[-1.2ex]{0pt}{0pt} \\ 
\hline
P$_{1}$ & 0.65 & 0.23 & 0.62 & 0.21 & 0.77 & 0.08 & 67.8$\pm$8.1 \rule{0pt}{2.4ex} \\
P$_{2}$ & 0.42 & 0.38 & 0.35 & 0.43 & 0.77 & 0.16 & 65.2$\pm$2.5 \\
P$_{3}$ & 0.51 & 0.29 & 0.64 & 0.22 & 0.82 & 0.12 & 74.5$\pm$4.6 \rule[-1.2ex]{0pt}{0pt} \\
\hline
\end{tabular}
\end{center}
\end{table}

\subsection{Defect inspection evaluation} 

\begin{figure}[t]
\centerline{\includegraphics[width=0.9\linewidth]{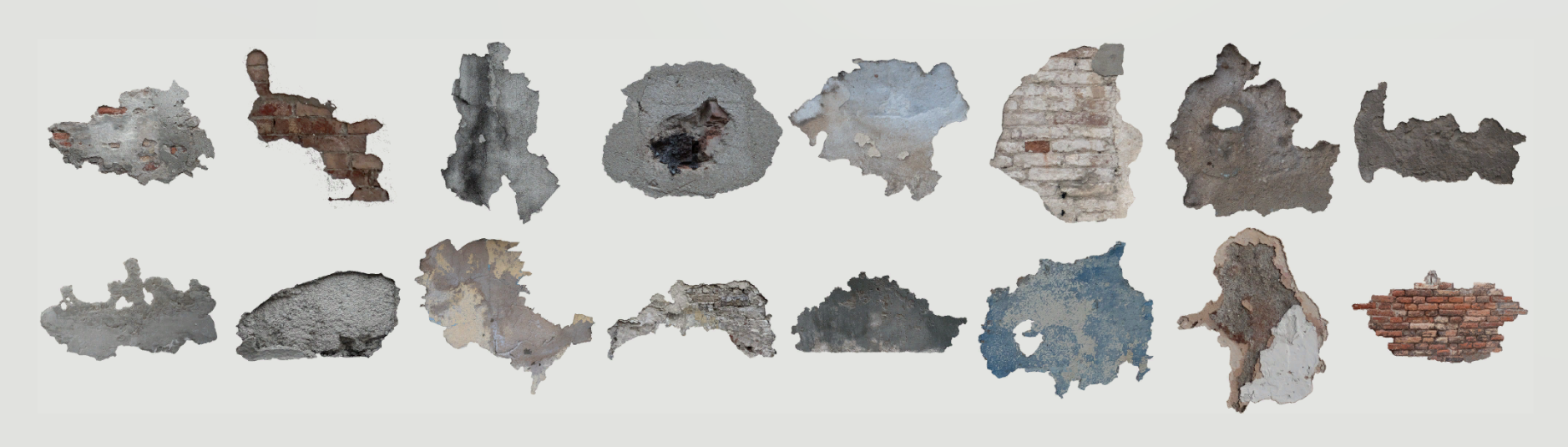}}
\caption{Samples of defects for defect inspection evaluation.}
\label{fig_exp2}
\end{figure}

To evaluate the defect size estimation algorithm and the estimate of drone's 5D pose, 16 different samples of defects were used (Fig. \ref{fig_exp2}). Each defect had two different sizes (small and large), so there were a total 32 defects to be tested. For small-size defects, the maximum length is less than 50 cm, with the range of area being 600--1400 $cm^2$. For large-size defects, the area is four times larger than the small-sized defects. The ground truth for the size of defects and the corresponding drone 5D pose were determined through simulation using Unity gaming engine by Unity Technologies. 

The drone's 2D orientation was not considered in the experiment as the size and the center of the defect is the major factor for the quality of the visual data collection of the drone. The drone's 3D position was split into 1D depth (z-axis) and 2D plane position (xy-plane). Depth is the distance from the defect in the surface normal direction, so it is related to the size of the defect on the image, while the plane position is the distance from the center of the defect.

Table \ref{table_exp3_result} shows the results of gaze-based defect evaluation, where $\Delta A$ is the error of the area of the defect, $\Delta d_{z}$ and $\Delta d_{xy}$ is the error of depth and plane position of the drone, respectively, and $t_{avg}$ is the average duration for data collection. Most of $\Delta A$ are below 10\% which is reasonable considering the results of the gaze accuracy test. $\Delta d_{z}$ and $\Delta d_{xy}$ are about 5\%, which makes only a marginal difference in the quality of the image taken from the drone position. These errors can be considered acceptable for visual infrastructure inspection purposes.

When participants inspected large-size defects, the overall accuracy was slightly increased or remained similar to the result of small-size defects. This is because the inaccuracy of gaze tracking has relatively less influence on defect estimation as the size of the defect is enough to overcome the error in gaze tracking. Furthermore, no significant differences were observed among participants. Based on these findings, it can be concluded that the performance of the defect size estimation remains consistent regardless of inspectors or the defect size (based on this limited study).

\begin{table}[t]
\caption{Results of defect inspection}
\label{table_exp3_result}
\begin{center}
\begin{tabular}{c c c c c c}
\hline
P & Size & $\Delta A $ (\%) & $\Delta d_{z}$ (\%) & $\Delta d_{xy}$ (\%) & $t_{avg}$ (sec) \rule{0pt}{2.4ex} \rule[-1.2ex]{0pt}{0pt} \\
\hline
\multirow{2}{*}{P$_{1}$} & S & 9.38$\pm$6.70 & 5.42$\pm$4.87 & 5.69$\pm$2.14 & 8.86$\pm$1.41 \rule{0pt}{2.4ex} \\ 
    & L & 5.26$\pm$4.15 & 3.71$\pm$2.84 & 2.84$\pm$1.29 & 10.5$\pm$1.22 \\ \cmidrule(lr){2-6}
\multirow{2}{*}{P$_{2}$} & S & 8.03$\pm$7.02 & 5.83$\pm$3.27 & 4.75$\pm$2.07 & 11.49$\pm$2.45  \\ 
    & L & 8.49$\pm$12.67 & 5.41$\pm$5.45 & 4.72$\pm$2.92 & 11.48$\pm$3.03 \\ \cmidrule(lr){2-6}
\multirow{2}{*}{P$_{3}$} & S & 8.35$\pm$12.07 & 6.30$\pm$8.53 & 4.80$\pm$3.49 & 10.46$\pm$1.86 \\ 
    & L & 3.99$\pm$4.53 & 5.41$\pm$4.71 & 3.67$\pm$2.05 & 10.86$\pm$1.85 \rule[-1.2ex]{0pt}{0pt}\\ 
\hline
\end{tabular}
\end{center}
\end{table}

\section{CONCLUSIONS}

In this study, a gaze-based HRI system for infrastructure inspection is proposed to enhance the capabilities of visual inspection experts using a head-mounted MR device. In order to exploit gaze as an effective input for defect inspections, human attention level is categorized into three steps (scanning, focusing, and inspecting) based on fixations and saccades. A visual cue in the form of an MR drone is displayed for situation awareness where an inspector can associate the algorithmic results with their intention, and correct their behavior if necessary. Once the human attention level meets the criteria of the inspecting step, the system saves all fixations until it collects a sufficient amount of data. The properties of the defect are evaluated based on the collected fixations and then the drone's 5D pose for photographing the defect is determined. Through several experiments, the utility of the gaze-based HRI system was demonstrated. 

The main conclusions of this study are as follows: (1) with an appropriate indicator for situational awareness, eye gaze can be utilized as a key input for the HRI system for infrastructure inspections; (2) human attention level can be tracked during the inspection activity through eye movements relatively non intrusively; and (3) experiments show that the gaze-based HRI system demonstrates sufficient accuracy and significant promise to aid in defect evaluation during routine infrastructure inspections.

In addition to the relatively small number of human participants, this study has additional limitations. Although meaningless eye movements can be discarded through our gaze pattern classification, distinguishing between intentional and unintentional eye movements during the inspection phase can be challenging. Additional methods may be employed to identify the intention to overcome this limitation but are not considered here. Furthermore, image information of defects was not utilized for the defect quantification in our approach but can be incorporated to increase the overall accuracy or robustness of the method.




\end{document}